\documentclass{article}
\usepackage{spconf,amsmath,graphicx,hyperref}
\usepackage{booktabs}  
\usepackage{stfloats}
\usepackage{amssymb} 
\usepackage{enumitem} 
\usepackage{multirow}

\title{Rethinking Self-Training Based Cross-Subject Domain Adaptation for SSVEP Classification}
\name{Weiguang Wang\textsuperscript{1}, Yong Liu\textsuperscript{1}, Yingjie Gao\textsuperscript{2,*}, Guangyuan Xu \textsuperscript{1,*} \thanks{\textsuperscript{*}Corresponding author:  Yingjie Gao (gaoyingjie@buaa.edu.cn), Guangyuan Xu (xuguangyuan@bupt.edu.cn)
}}
\address{Author Affiliation(s)}

\address{\textsuperscript{1} School of Artificial Intelligence, Beijing University of Posts and Telecommunications, Beijing, China \\
	\textsuperscript{2} School of Computer Science and Technology, Beihang University, Beijing, China \\
	}

\begin{document}
\ninept

\maketitle
\begin{abstract}
\begin{sloppypar}
Steady-state visually evoked potentials (SSVEP)-based brain–computer interfaces (BCIs) are widely used due to their high signal-to-noise ratio and user-friendliness. Accurate decoding of SSVEP signals is crucial for interpreting user intentions in BCI applications. However, signal variability across subjects and the costly user-specific annotation limit recognition performance. Therefore, we propose a novel cross-subject domain adaptation method built upon the self-training paradigm. 
Specifically, a Filter-Bank Euclidean Alignment (FBEA) strategy is designed to exploit frequency information from SSVEP filter banks. 
Then, we propose a Cross-Subject Self-Training (CSST) framework consisting of two stages: Pre-Training with Adversarial Learning (PTAL), which aligns the source and target distributions, and Dual-Ensemble Self-Training (DEST), which refines pseudo-label quality.
Moreover, we introduce a Time-Frequency Augmented Contrastive Learning (TFA-CL) module to enhance feature discriminability across multiple augmented views. Extensive experiments on the Benchmark and BETA datasets demonstrate that our approach achieves state-of-the-art performance across varying signal lengths, highlighting its superiority.
\end{sloppypar}
\end{abstract}
\begin{keywords}
SSVEP, domain adaptation, adversarial learning, self-training, contrastive learning
\end{keywords}
\section{Introduction}
\label{sec:intro}
Brain–Computer Interfaces (BCIs) construct a direct bridge between the brain and external devices, facilitating information exchange without reliance on traditional neural or muscular pathways~\cite{BCI2025}. By analyzing electroencephalogram (EEG) signals, user intentions can be decoded, thereby enabling advanced applications in communication and human–machine interaction~\cite{EEG}. Among  EEG paradigms, steady-state visual evoked potential (SSVEP) is widely used due to its high signal-to-noise ratio and user-friendliness~\cite{al2024comprehensive}. It has been applied in tasks such as spelling and robotic control~\cite{mai2024hybrid}, demonstrating great potential in information transfer. 


Accurate SSVEP decoding is essential for SSVEP-based BCIs~\cite{ding2025eeg}. In early studies, training-free methods such as CCA~\cite{CCA} and FBCCA~\cite{FBCCA} are proposed, which compute the correlation coefficients between the recorded EEG and the reference. However, their dependence on fixed reference signals leads to poor generalization in cross-subject scenarios. To alleviate this issue, domain generalization based methods propose to form templates or train models using source subjects, and then directly apply them to the target subject.
tt-CCA~\cite{ttCCA} forms source-based templates to reduce the discrepancy between the templates and target subjects.
Ensemble-DNN~\cite{Ensemble-DNNs} constructs subject-specific models on the source domain and combines the predictions from the $k$ models most similar to a new user, thereby improving robustness.
Nevertheless, as these methods do not utilize target data for adaptation, their performance remains limited.


Domain adaptation based approaches provide a promising solution for cross-subject SSVEP-based BCIs, as they adapt to the target domain in an unsupervised manner, further enhancing classification performance.
OACCA~\cite{OACCA} develops an online adaptation method that updates spatial filters using online unlabeled data from the target domain. SUTL~\cite{SUTL} selects suitable transferable source subjects and aligns these selected subjects with the target subject, thereby improving recognition accuracy for the new subject.


Recently, self-training based methods have emerged as an effective strategy for unsupervised domain adaptation.
Specifically, the model is first trained on labeled source data, then used to generate pseudo-labels for target domain, and the model is updated using these pseudo-labels. 
SFDA~\cite{SFDA} is the first to introduce the self-training paradigm to cross-subject SSVEP classification, which designs a local regularization term to adaptively update neighboring labels and improve pseudo-label quality. 
However, merely improving pseudo-label quality through post-processing is suboptimal.

In this work, we first propose a Filter-Bank Euclidean Alignment (FBEA) strategy tailored for SSVEP signals, thereby enabling better exploitation of complementary information across frequencies.
Then, we design a novel Cross-Subject Self-Training (CSST) framework.
In the pre-training stage, adversarial learning is employed to align the source and target distributions, thereby improving model generalization.
In the self-training stage, we propose Dual-Ensemble Self-Training (DEST), where a temporally-ensembled teacher model generates multi-view ensemble pseudo-labels to improve pseudo-label quality.
Finally, since self-training still suffers from noisy labels, we introduce a Time–Frequency Augmented Contrastive Learning (TFA-CL) module to enhance feature discrimination.

In summary, our contributions are as follows:
\begin{itemize}[itemsep=0pt, topsep=1pt, parsep=0pt, partopsep=0pt]
    \item We propose a Filter-Bank Euclidean Alignment strategy that aligns SSVEP frequency bands to better capture their harmonic components. 
    \item We design a Cross-Subject Self-Training framework, which first applies domain adversarial learning to align source and target distributions, and subsequently adopts a dual-ensemble mechanism to refine pseudo-label quality.
    \item We introduce a Time-Frequency Augmented Contrastive Learning module, which learns discriminative features across multiple augmented views.
    \item Extensive experiments are conducted on two major datasets, where our method achieves state-of-the-art performance across varying signal lengths, highlighting its effectiveness.
\end{itemize}

\begin{figure*}[!t]
  \centering
  \includegraphics[width=0.82\textwidth]{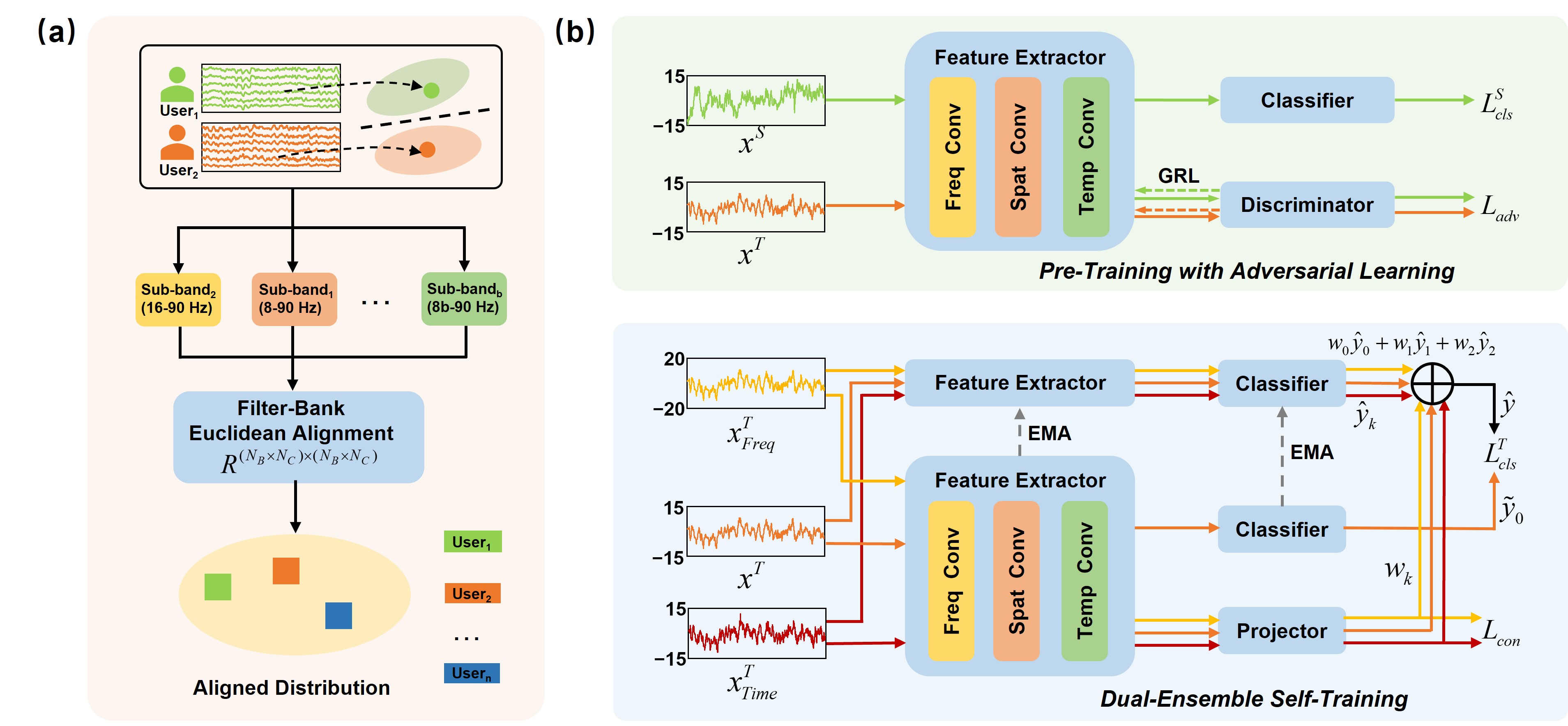}
  \caption{Overview of the proposed cross-subject domain adaptation method. (a) The strategy of Filter-Bank Euclidean Alignment. (b) The framework of Cross-Subject Self-Training consisting of a Pre-Training with Adversarial Learning stage (top) and a Dual-Ensemble Self-Training stage (bottom), where Time-Frequency Contrastive Learning is applied in the second stage.}
  \label{fig:cross}
\end{figure*}

\section{Methodology}
\label{sec:format}
In cross-subject SSVEP-based BCIs, we are given a labeled source domain $\mathcal{D}_S = \{(x_i^s, y_i^s)\}_{i=1}^{n_s}$ consisting of $n_s$ EEG samples, where $x_i^s \in \mathbb{R}^{N_C \times N_P}$ denotes the $i$-th EEG sample with $N_C$ channels and $N_P$ sampling points, and $y_i^s$ denotes its corresponding label.
In addition, we have access to an unlabeled target domain $\mathcal{D}_T = \{x_i^t\}_{i=1}^{n_t}$. Due to magnitude variability across subjects, the source and target domains are assumed to follow different distributions (\emph{i.e.}, $P_S(x) \neq P_T(x)$).
The goal is to improve the classification performance in $\mathcal{D}_T$ using the knowledge in $\mathcal{D}_S$.

As shown in Fig.~\ref{fig:cross}(a), Filter-Bank Euclidean Alignment (FBEA) is first proposed to reduce the marginal distribution shift between the source and target domains.
We then introduce a novel Cross-Subject Self-Training (CSST) framework, illustrated in Fig.~\ref{fig:cross}(b). It consists of two stages: a Pre-Training with Adversarial Learning (PTAL) stage and a Dual-Ensemble Self-Training (DEST) stage. 
Furthermore, Time–Frequency Augmented Contrastive Learning (TFA-CL) is applied to the DEST stage to enhance feature discrimination.
\subsection{Filter-Bank Euclidean Alignment}
\label{ssec:subhead}
Considering the magnitude variability and inherent noise of EEG signals across subjects, data alignment is commonly employed to mitigate the marginal distribution shift between $P_S(x)$ and $P_T(x)$.
Previous preprocessing strategies, such as channel normalization and trial normalization \cite{xu2021enhancing, li2025inter}, are restricted to eliminating amplitude differences and fail to  align the data distributions across subjects. Conventional euclidean alignment \cite{EU_wu2025revisiting} typically operates at the channel level by matching the covariance matrices derived from EEG data. 
However, SSVEP signals contain non-negligible harmonic responses, which require filter banks to capture information from multiple frequency bands. Channel euclidean alignment may therefore overlook the complementary information across frequencies. To overcome this limitation, we propose an FBEA strategy.

Specifically, the signal after filter-bank decomposition is denoted as  $x \in \mathbb{R}^{N_B \times N_C \times N_P}$, where $N_B$ is the number of filter banks. The reference matrix \(\bar{R}\) is defined as: 
\begin{equation}\bar{R} = \frac{1}{n} \sum_{i=1}^n x_i x_i^T,\end{equation}
where $\bar{R} \in \mathbb{R}^{(N_B \times N_C) \times (N_B \times N_C)}$ denotes the mean covariance matrix across all samples.
Alignment is then performed as \begin{equation}\tilde{x}_i = \bar{R}^{-1/2} x_i.
\end{equation}
After alignment, the covariance matrix becomes the identity matrix, thereby achieving signal whitening.  The proposed FBEA strategy leverages filter-bank frequency information to realize more precise alignment, and effectively reduces distribution shift.

\subsection{Cross-Subject Self-Training}
\label{ssec:subhead}
EEG signals are inherently non-stationary and exhibit substantial variability across subjects. Recent studies~\cite{Ensemble-DNNs, SFDA} have introduced the self-training (ST) paradigm to boost performance on target domains. The ST paradigm involves two main stages. In the source domain pre-training stage, we conduct supervised training using annotated source EEG data to obtain an initial model. During the target domain self-training stage, the pre-trained model generates pseudo-labels for the target domain, which are then used to update the model.

Although ST based cross-subject methods have achieved remarkable progress, they still face the issue of low-quality pseudo-labels. We propose a novel Cross-Subject Self-Training framework, which improves both stages individually.

\textbf{Pre-Training with Adversarial Learning.}
As a classic approach to unsupervised domain adaptation, adversarial learning~\cite{xiao2024improved} can reduce the domain gap by aligning the feature distributions of the source and target domains. 
To this end, we incorporate adversarial learning in the source domain pre-training stage to learn discriminative feature representations, thereby improving generalization to the target domain and yielding higher-quality pseudo-labels. 

Given that SSVEP signals exhibit spectral, spatial, and temporal characteristics, we employ a CNN-based network  $G$ \cite{Ensemble-DNNs} for feature representation, which integrates filter-bank fusion, spatial filtering, and temporal feature extraction, along with a classification head $H$. To align the feature distributions across domains, a domain classifier $D$ is introduced. $G$ and $D$ are connected via the Gradient Reverse Layer (GRL)~\cite{ganin2016domain}, which reverses the gradients that flow through $G$.
The loss function in the first stage is defined as follows:
\begin{equation}
L_{\mathrm{cls}}^{S}
= \mathbb{E}_{(x,y)\sim \mathcal{D}_S} 
\mathrm{CE}\big(H(G(x)),y\big),
\label{eq:dann_cls}
\end{equation}
\begin{equation}
\begin{split}
L_{\mathrm{adv}} 
= \max_{\theta_G} \min_{\theta_D}  
     \mathbb{E}_{x_s \sim \mathcal{D}_S} \log D(G(x_s)) \\
    + \mathbb{E}_{x_t \sim \mathcal{D}_T} \log (1 - D(G(x_t)) ),
\end{split}
\end{equation} 
\begin{equation}
L_{\mathrm{pre-training}}
= L_{\mathrm{cls}}^{S}
+  L_{\mathrm{adv}}, 
\label{eq:dann_first}
\end{equation}

\textbf{Dual-Ensemble Self-Training.}
Traditional self-training based methods~\cite{weng2025self, rafiei2022self} are inevitably affected by low-quality pseudo-labels, and directly using them causes error accumulation. To address this, we propose Dual-Ensemble Self-Training, which first leverages the mean-teacher paradigm to iteratively refine unlabeled target data. This paradigm employs two identical models: a teacher and a student. Specifically, the teacher first makes predictions on the target data and generates pseudo-labels through confidence-based filtering, which are then used to guide the student’s learning. The teacher's parameters $\theta_t$ are updated as the exponential moving average (EMA) of the student's parameters $\theta_s$, defined as: 
\begin{equation}
\theta_t = \alpha \, \theta_t + (1-\alpha)\,\theta_s,
\label{eq:ema_update}
\end{equation}
where $\alpha$ is the EMA momentum coefficient. Thus, the teacher model acts as a temporal ensemble of the student.

On the other hand, we design a pseudo-label fusion strategy, serving as an ensemble over multiple augmented views. Let $\hat{y}_0$ denote the pseudo-label of the original data, $\hat{y}_1, \hat{y}_2$ denote the pseudo-labels obtained from two augmented views. The fused pseudo-label is defined as:
\begin{equation}
\hat{y} \;=\; \sum_{k=0}^{2} w_k \,\hat{y}_k,
\label{eq:pseudo_fusion}
\end{equation}
\begin{equation}
w_k \;=\; \frac{\exp\!\big( \cos(z_k, z_0)\big)}
{\sum_{m=0}^{2} \exp\!\big(\cos(z_m, z_0)\big) + \epsilon},
\label{eq:pseudo_fusion_weight}
\end{equation}
where $z_k = P(G(x_k))$ represents the projected feature, $\cos(\cdot,\cdot)$ denotes the cosine similarity computed with respect to the original-view projection $z_0$, and $\epsilon$ is set to $10^{-8}$ for numerical stability.

\subsection{Time–Frequency Augmented Contrastive Learning}
\label{ssec:subhead}
Although obtaining reliable pseudo-labels substantially improves performance, self-training based on cross-entropy still suffers from noisy labels. To address this, we propose a Time-Frequency Augmented Contrastive Learning module to enhance feature discrimination.
For SSVEP signals, we introduce two types of data augmentation: temporal perturbation and noise injection. These augmentations operate along the temporal and spectral dimensions to generate diverse views. Based on these views, we employ the supervised contrastive loss \cite{dai2025contrastive}, which encourages representations of the same class to cluster together while pushing apart those of different classes. Formally, the contrastive loss is defined as:
\begin{equation}
{L}^i_{\mathrm{con}}
= - \frac{1}{|pos(i)|}
  \sum_{p \in pos(i)}
  \log
  \frac{\exp\left( z_i^\top z_p / {\tau} \right)}
       {\sum\limits_{a \in A \setminus \{i\}}
        \exp\left( z_i^\top z_a / {\tau} \right)}
,
\end{equation}
where $pos(i)$ denotes the augmented samples that share the same predicted class as $x_i$, $A$ denotes all augmented samples in the batch, and $\tau$ is the temperature parameter.
The loss function in the second stage is defined as follows:
\begin{equation}
L_{\mathrm{self-training}}
= L_{\mathrm{cls}}^{T}
+  \lambda_{con} L_{\mathrm{con}}, 
\label{eq:dann_first}
\end{equation}


\section{Experiments}
\label{sec:pagestyle}
\subsection{Dataset and Experimental Setup}
\textbf{Dataset.}
We evaluate our method on two large-scale SSVEP datasets, Benchmark \cite{wang2016benchmark} and BETA \cite{liu2020beta}, which provide 64-channel EEG recordings from 35 and 70 subjects performing a BCI spelling task. Both datasets comprise 40 target characters with stimulation frequencies ranging from 8 to 15.8 Hz in 0.2 Hz steps, and a phase difference of $0.5\pi$ between adjacent frequencies. In the Benchmark dataset, each subject completed six blocks with a stimulus duration of 5 s. In the BETA dataset, each subject completed four blocks; the first 15 subjects had a stimulus duration of 2 s, while the remaining 55 subjects had a duration of 3 s.

In this study, the EEG signals were preprocessed as follows:  1) 9 channels over the occipital region were selected (\emph{i.e.}, Pz, PO5, PO3, POz, PO4, PO6, O1, Oz and O2); 2) The signals were segmented into intervals $[T_d, T_d + T_w]$, where $T_d$ denotes the corresponding stimulus latency (0.14 s / 0.13 s), and $T_w$ represents the data length; 3) Filter banks were applied to decompose the EEG signals into $N_B$ sub-bands.

\textbf{Implementation Details.}
We optimize the model using the Adam algorithm with a learning rate of 0.0001 and a weight decay of 0.001. The batch size is 64, and the number of epochs is set to 500 for the first stage and 500 for the second stage. During training, cross-entropy loss is applied to both classification and domain discrimination. The pseudo-label threshold is set at 0.9, and the momentum coefficient of EMA is set to 0.999. For contrastive learning, the temperature parameter is set to $\tau = 0.5$, and the loss weight is $\lambda_{\text{con}} = 0.01$.

\textbf{Performance Evaluation.}
We compare our method against five approaches, including the domain generalization method tt-CCA~\cite{ttCCA}, Ensemble-DNN~\cite{Ensemble-DNNs}, the domain adaptation methods OACCA~\cite{OACCA}, SUTL~\cite{SUTL}, and SFDA~\cite{SFDA}. 
Our method follows the Leave-One-Subject-Out (LOSO) principle for cross-subject evaluation, ensuring consistency with the comparison methods. 

The performance is evaluated using accuracy and information transfer rate (ITR), with ITR defined as:  
\begin{equation}
\mathrm{ITR} = \frac{60}{T} \times 
\left[ \log_2 M 
+ P\log_2 P 
+ (1 - P)\log_2 \left( \frac{1 - P}{M - 1} \right) \right],
\label{eq:ITR}
\end{equation}
where $M$ is the number of targets, $P$ is the classification accuracy, and $T$ denotes the total time (including both gaze and shift time).

\begin{figure}[t]
  \centering
  \includegraphics[width=\linewidth]{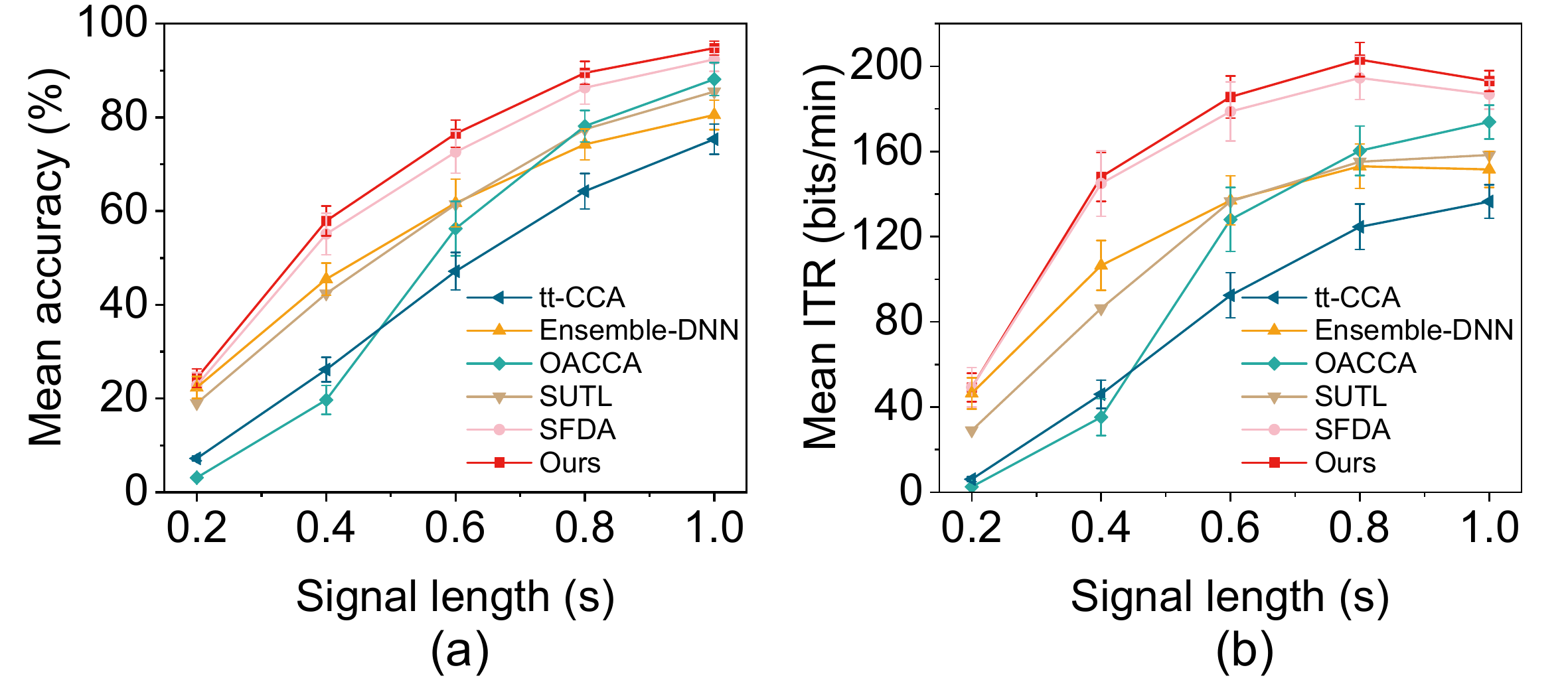}
  \caption{Comparisons of classification accuracy (a) and ITR (b) among six different methods  on the Benchmark dataset. Standard errors are indicated by the bars.}
  \label{fig:fig2}
\end{figure}
\vspace{-1.2em} 

\subsection{Results and Comparisons}
\label{ssec:subhead}
\begin{figure}[!t]
  \centering
  \includegraphics[width=\linewidth]{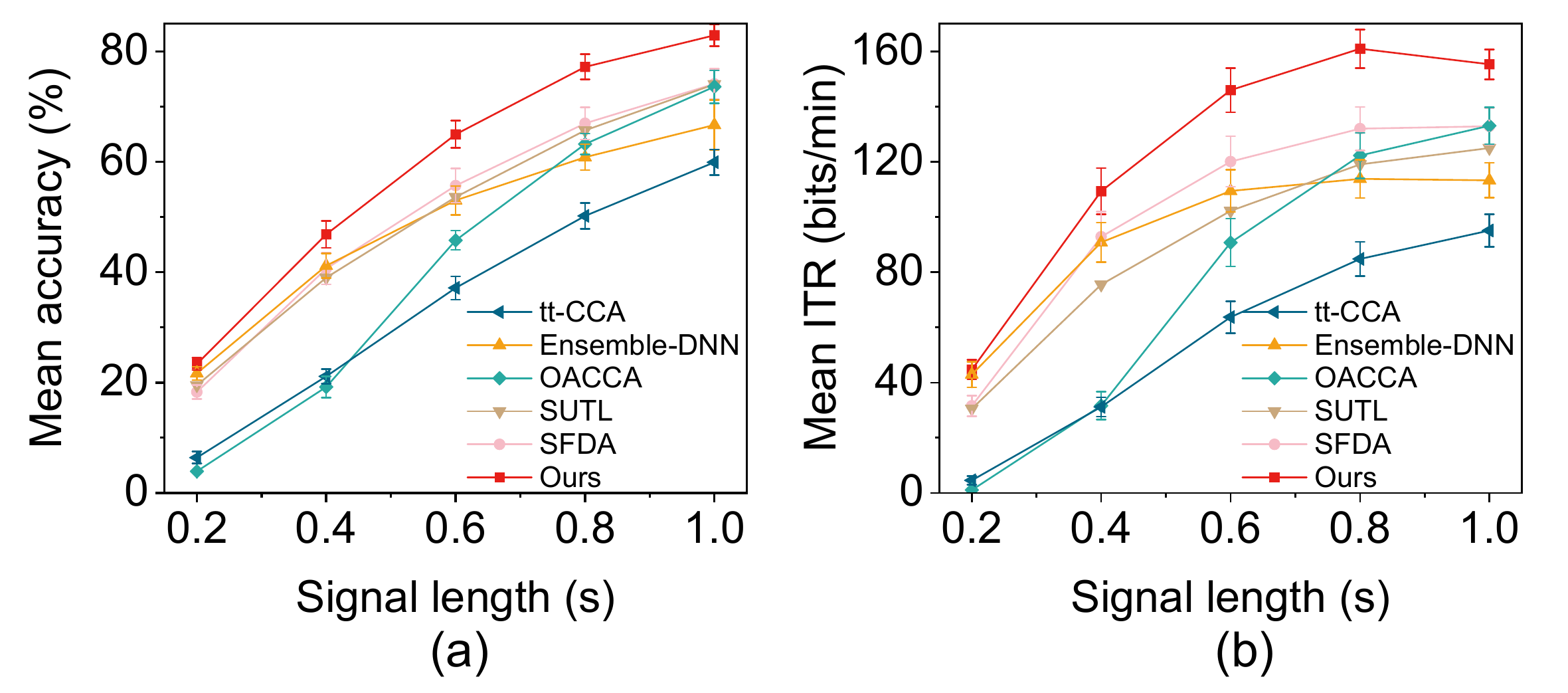}
  \caption{Comparisons of classification accuracy (a) and ITR (b) among six different methods  on the BETA dataset. Standard errors are indicated by the bars.}
  \label{fig:fig3}
\end{figure}

As shown in Fig.~\ref{fig:fig2} and Fig.~\ref{fig:fig3}, the classification performance of various methods at different signal lengths is presented. Our method consistently demonstrates superior performance in both classification accuracy and ITR across all signal lengths on the two datasets.
For the Benchmark dataset, our method yields the highest ITR of $203.1 \pm 8.03$ bits/min at $0.8$ s, significantly surpassing the latest SOTA method SFDA, which achieves $194.54 \pm 10.07$ bits/min ($p < 0.01$).
For the BETA dataset, the highest ITR of $160.93 \pm 6.93$ bits/min is achieved at $0.8$\,s, which is significantly higher than that of SFDA ($131.99 \pm 7.86$ bits/min; $p < 0.001$)).
These results provide compelling evidence for the effectiveness of our proposed method in
addressing cross-subject domain adaptation challenges in SSVEP-based BCI tasks.

\subsection{Ablation Study}
\textbf{Effectiveness of each component.}
To verify the effectiveness of each component, we conduct ablation experiments on the Benchmark dataset with the signal length of 1 s, as shown in Tab.~\ref{tab:ablation}. 
Row 1 shows the baseline using pure self-training, where supervised training is performed on the source data and self-training with pseudo-labels on the target data, achieving $85.13\%$ accuracy.
Based on this, after adopting our PTAL in the first stage, the accuracy improves by $7.22\%$, and it further improves by $1.07\%$ through incorporating our DEST in the second stage. 
This improvement is attributed to adversarial learning for better generalization to the target domain and the dual ensemble that refines pseudo-labels across temporal and multiple views.
Additionally, our FBEA yields improvements on both the baseline ($0.95\%$) and  CSST ($0.94\%$). TFA-CL further enhances the performance by learning more robust representations, achieving $94.80\%$ accuracy.
    
\begin{table}[!h]
  \centering
  \caption{Ablation study of each component on the Benchmark dataset with a data length of 1 s.}
  \vspace{0.2cm}
  \label{tab:ablation}
 
  \begin{tabular}{@{}cccccc@{}}  
    \toprule
    \multirow{2}{*}{FBEA} & \multicolumn{2}{c}{CSST} & \multirow{2}{*}{TFA-CL} & \multirow{2}{*}{Acc (\%)} & \multirow{2}{*}{ITR (bits/min)} \\
    \cmidrule(lr){2-3} 
    & PTAL & DEST &  &  \\
    \midrule
     & & & & 85.13 & 166.05\\ 
    \midrule
    \checkmark & & & & 86.08 & 169.77 \\
     & \checkmark & & & 92.35 & 185.51\\
     & \checkmark & \checkmark & & 93.42 & 188.97\\
     \checkmark & \checkmark & \checkmark & & 94.36 & 191.87\\
     \checkmark & \checkmark & \checkmark & \checkmark & \textbf{94.80} & \textbf{193.12}\\
    
    \bottomrule
  \end{tabular}
 
\end{table}
\textbf{Detailed analysis of CSST.}
We compare our proposed CSST framework with the baseline pure self-training method on the Benchmark dataset with a data length of 1 s. As shown in Tab.~\ref{tab:two stages}, Self-training achieves $77.07\%$ accuracy on the target domain after supervised training on labeled source EEG data.
Using pseudo-labels for fine-tuning on the target domain data increases the accuracy to $85.13\%$.
In contrast, our CSST framework leverages the target domain data for adversarial learning in the first stage, which enhances the model's generalization and achieves $88.36\%$ accuracy, even surpassing the result of the baseline's second stage. 
With DEST integrated, classification accuracy reaches $93.42\%$, demonstrating the dual ensemble’s effectiveness in enhancing  the quality of pseudo-labels.

\begin{table}[!h]
  \centering
  \caption{Detailed analysis of CSST on the Benchmark dataset with a data length of 1 s.}  
  \vspace{0.1cm}
  \begin{tabular}{ccc}  
    \toprule  
    Method       & First Stage & Second Stage \\
    \midrule  
    Self-Training           &     77.07      &   85.13   \\
    CSST         &    \textbf{88.36}       &  \textbf{93.42}   \\
    \bottomrule  
  \end{tabular}
  \label{tab:two stages}  
\end{table}

\textbf{Ablation on FBEA.}
As summarized in Tab.~\ref{tab:alignment}, different preprocessing methods yield varying performance. 
Channel normalization and trial normalization achieve accuracies of $93.87\%$ and $93.58\%$, respectively, both slightly lower than the $93.96\%$ obtained without any preprocessing. This is because these methods fail to address inter-subject covariate shift, and unreasonable normalization may even weaken the model's generalization.
In contrast, euclidean alignment boosts accuracy by aligning the geometric centers of the source and target domains. However, the channel euclidean alignment operates solely on the spatial dimension and overlooks the frequency information from SSVEP filter banks. By incorporating this complementary information, FBEA achieves the highest classification accuracy, demonstrating its superiority.

\begin{table}[!h]
  \centering
  \caption{Comparison of different preprocessing strategies on the Benchmark dataset with a data length of 1 s.}
  \setlength{\tabcolsep}{5pt} 
  \vspace{0.1cm}
  \begin{tabular}{lccccc}
    \toprule
    - & Channel Norm & Trial Norm & Channel Euclid & FBEA \\
    \midrule
    93.96 & 93.87 & 93.58 & 94.25 & \textbf{94.80} \\
    \bottomrule
  \end{tabular}
  \label{tab:alignment}
\end{table}

\section{Conclusion}
\label{sec:typestyle}
In this study, we propose a cross-subject domain adaptation method for SSVEP-based BCIs. First, the SSVEP signals are aligned through an FBEA strategy to capture complementary frequency information. Then, we design a CSST framework, where PTAL aligns source and target distributions, and DEST refines pseudo-labels across temporal and multiple views. Furthermore, we introduce a TFA-CL module to enhance feature discriminability across multiple augmented views. 
Experimental results show that our method achieves superior performance in both classification accuracy and ITR across varying signal lengths on the two datasets. 
In the future, our method is expected to support practical cross-subject SSVEP-based BCI applications in daily life. 




\vfill\pagebreak
\section{Acknowledgement}
This work was supported by the National Science and Technology Major Project Fund of China (Grant No.~2025ZD0215600), the General Program of the Natural Science Foundation of Beijing (Grant No.~7242271), the Shanghai Special Fund for Industrial High-Quality Development: Pioneering Industries Innovation and Development Projects (Grant No.~2024-GZL-RGZN-02026), the Start-up Funding of Beijing University of Posts and Telecommunications (Grant No.~500422817 and 510224039), and the Beijing University of Posts and Telecommunications Excellent Ph.D. Students Foundation (Grant No.~CX20241089).

\section{Compliance with Ethical Standards}
This  work used public datasets released in~\cite{wang2016benchmark, liu2020beta}. Ethical approval was not required as confirmed by the license attached with the open access data.




\bibliographystyle{IEEEbib}
\bibliography{template}

\end{document}